\documentclass[11pt,a4paper]{article}
\usepackage[hyperref]{acl2020}
\usepackage{times}
\usepackage{latexsym}

\usepackage{multirow}
\usepackage{tabularx}
\usepackage{amsmath}
\usepackage{wrapfig}
\usepackage{enumitem}
\usepackage{array}
\usepackage{wrapfig}
\usepackage[utf8]{inputenc}
\newcolumntype{L}{>{\raggedright\arraybackslash}m{0.97\linewidth}}
\usepackage{graphicx}       
\usepackage{booktabs}       
\usepackage{amsfonts}       
\usepackage{nicefrac}       
\usepackage{microtype}      
\usepackage{float,lscape}
\usepackage{xcolor}
\usepackage{tikz}
\usepackage{color, colortbl}

\usepackage{soul}
\usepackage{latexsym}
\usetikzlibrary{calc}
\usetikzlibrary{decorations.pathmorphing}
\definecolor{myblue}{RGB}{204, 229, 255}
\definecolor{myred}{RGB}{255, 205, 205}
\definecolor{myyellow}{RGB}{253, 253, 153}
\definecolor{mygreen}{RGB}{179, 253, 179}
\DeclareRobustCommand{\hlred}[1]{{\sethlcolor{myred}\hl{#1}}}
\DeclareRobustCommand{\hlblue}[1]{{\sethlcolor{myblue}\hl{#1}}}
\DeclareRobustCommand{\hlyellow}[1]{{\sethlcolor{myyellow}\hl{#1}}}
\DeclareRobustCommand{\hlviolet}[1]{{\sethlcolor{mygreen}\hl{#1}}}

\aclfinalcopy 
\def\aclpaperid{1844} 

\title{Generating Fact Checking Explanations}

\author{Pepa Atanasova \text{    } Jakob Grue Simonsen \text{    } Christina Lioma \text{    } Isabelle Augenstein\\
Department of Computer Science \\
University of Copenhagen \\
Denmark \\
\texttt{\{pepa, simonsen, c.lioma, augenstein\}@di.ku.dk} \\}

\date{}

\begin{document}
\maketitle
\begin{abstract}
Most existing work on automated fact checking is concerned with predicting the veracity of claims based on metadata, social network spread, language used in claims, and, more recently, evidence supporting or denying claims. A crucial piece of the puzzle that is still missing is to understand how to automate the most elaborate part of the process -- generating justifications for verdicts on claims.
This paper provides the first study of how these explanations can be generated automatically based on available claim context, and how this task can be modeled jointly with veracity prediction. Our results indicate that optimising both objectives at the same time, rather than training them separately, improves the performance of a fact checking system. The results of a manual evaluation further suggest that the informativeness, coverage and overall quality of the generated explanations are also improved in the multi-task model.


\end{abstract}

\section{Introduction}\label{sec:intro}
\noindent When a potentially viral news item is rapidly or indiscriminately published by a news outlet, the responsibility of verifying the truthfulness of the item is often passed on to the audience. To alleviate this problem, independent teams of professional fact checkers manually verify the veracity and credibility of common or particularly check-worthy statements circulating the web. However, these teams have limited resources to perform manual fact checks, thus creating a need for automating the fact checking process.
\setlength{\tabcolsep}{0.2em}

\begin{table}[t!]
\fontsize{8.4}{8.4}\selectfont
\begin{center}
\begin{tabular}{|L|}
\toprule
\textbf{Claim}: The last major oil spill from a drilling accident in America happened over 40 years ago in 1969.\\ \midrule
\textbf{Ruling Comments}: 
(...) \hlyellow{The last major oil spill from a drilling accident in America happened over 40 years ago in 1969.} \\
\hspace*{3mm}(...) The largest in volume was the Santa Barbara spill of 1969 referenced by Murdock and Johnson, in which an estimated 100,000 barrels of oil spilled into the Pacific Ocean, according to the API. \hlblue{The Santa Barbara spill was so big it ranked seventh among the 10 largest oil spills caused by marine well blowouts in the world, the report states.} Two other U.S. spills, both in 1970, rank eighth and 10th. \hlred{Fourteen marine blowouts have taken place in the U.S. between 1969 and 2007.} Six of them took place after 1990 and spilled a total of nearly 13,700 barrels. \\
\hspace*{3mm}(...) We interviewed three scientists who said that the impact of a spill has little to do with its volume. \hlviolet{Scientists have proven that spills far smaller than Santa Barbara's have been devastating.} \\  \midrule
\textbf{Justification}: While the nation's largest oil well blowout did take place in 1969, it's not factually correct to call it the ``last major oil spill". First of all, two of the largest blowouts in the world took place in the U. S.  the following year. More importantly, experts agree that spills far smaller in volume to the 1969 disaster have been devastating. From a scientific perspective, Johnson's decision to single out the 1969 blowout as the last ``major" one makes no sense. \\ \midrule

\textbf{Ruling}: Half-True \\ \bottomrule
\end{tabular}
\end{center}
\caption{\label{tab:Example} Example instance from the LIAR-PLUS dataset, with oracle sentences for generating the justification highlighted.}
\end{table}

The current research landscape in automated fact checking is comprised of systems that estimate the veracity of claims based on available metadata and evidence pages. Datasets like LIAR \cite{wang2017liar} and the multi-domain dataset MultiFC \cite{augenstein-etal-2019-multifc} provide real-world benchmarks for evaluation. There are also artificial datasets of a larger scale, e.g., the FEVER \cite{Thorne18Fever} dataset based on Wikipedia articles. As evident from the effectiveness of state-of-the-art methods for both real-world -- 0.492 macro F1 score \cite{augenstein-etal-2019-multifc}, and artificial data -- 68.46 FEVER score (label accuracy conditioned on evidence provided for `supported' and `refuted' claims) \cite{stammbach-neumann-2019-team},  the task of automating fact checking remains a significant and poignant research challenge.

A prevalent component of existing fact checking systems is a stance detection or textual entailment model that predicts whether a piece of evidence contradicts or supports a claim \cite{Ma:2018:DRS:3184558.3188729, mohtarami-etal-2018-automatic, Xu2019AdversarialDA}. Existing research, however, rarely attempts to directly optimise the selection of relevant evidence, i.e., the self-sufficient explanation for predicting the veracity label \cite{Thorne18Fever, stammbach-neumann-2019-team}.
On the other hand, \citet{alhindi-etal-2018-evidence} have reported a significant performance improvement of over 10\% macro F1 score when the system is provided with a short human explanation of the veracity label. Still, there are no attempts at automatically producing explanations, and automating the most elaborate part of the process - producing the \emph{justification} for the veracity prediction - is an understudied problem.

In the field of NLP as a whole, both explainability and interpretability methods have gained importance recently, because most state-of-the-art models are large, neural black-box models. Interpretability, on one hand, provides an overview of the inner workings of a trained model such that a user could, in principle, follow the same reasoning to come up with predictions for new instances. However, with the increasing number of neural units in published state-of-the-art models, it becomes infeasible for users to track all decisions being made by the models.
Explainability, on the other hand, deals with providing local explanations about single data points that suggest the most salient areas from the input or are generated textual explanations for a particular prediction.

Saliency explanations have been studied extensively \cite{Adebayo:2018:SCS:3327546.3327621, arras-etal-2019-evaluating, poerner-etal-2018-evaluating}, however, they only uncover regions with high contributions for the final prediction, while the reasoning process still remains behind the scenes. An alternative method explored in this paper is to generate textual explanations. In one of the few prior studies on this, the authors find that feeding generated explanations about multiple choice question answers to the answer predicting system improved QA performance \cite{rajani-etal-2019-explain}.

Inspired by this, we research how to generate explanations for veracity prediction. We frame this as a summarisation task, where, provided with elaborate fact checking reports, later referred to as \textit{ruling comments}, the model has to generate \textit{veracity explanations} close to the human justifications as in the example in Table~\ref{tab:Example}. We then explore the benefits of training a joint model that learns to generate veracity explanations while also predicting the veracity of a claim.\\
In summary, our \textbf{contributions} are as follows:
\begin{enumerate}[noitemsep]
\item{We present the first study on generating veracity explanations, showing that they can successfully describe the reasons behind a veracity prediction.}
\item{We find that the performance of a veracity classification system can leverage information from the elaborate ruling comments, and can be further improved by training veracity prediction and veracity explanation jointly.}
\item{We show that optimising the joint objective of veracity prediction and veracity explanation produces explanations that achieve better coverage and overall quality and serve better at explaining the correct veracity label than explanations learned solely to mimic human justifications.}
\end{enumerate}

\section{Dataset}\label{sec:dataset}

Existing fact checking websites publish claim veracity verdicts along with ruling comments to support the verdicts. Most ruling comments span over long pages and contain redundancies, making them hard to follow. Textual explanations, by contrast, are succinct and provide the main arguments behind the decision. PolitiFact~\footnote{https://www.politifact.com/} provides a summary of a claim's ruling comments that summarises the whole explanation in just a few sentences. 

We use the PolitiFact-based dataset LIAR-PLUS \cite{alhindi-etal-2018-evidence}, which contains 12,836 statements with their veracity justifications. The justifications are automatically extracted from the long ruling comments, as their location is clearly indicated at the end of the ruling comments. Any sentences with words indicating the label, which \citet{alhindi-etal-2018-evidence} select to be identical or similar to the label, are removed. We follow the same procedure to also extract the ruling comments without the summary at hand.

We remove instances that contain fewer than three sentences in the ruling comments as they indicate short veracity reports, where no summary is present. The final dataset consists of 10,146 training, 1,278 validation, and 1,255 test data points. A claim's ruling comments in the dataset span over 39 sentences or 904 words on average, while the justification fits in four sentences or 89 words on average. 

\section{Method}\label{sec:method}
We now describe the models we employ for training separately (1) an explanation extraction and (2) veracity prediction, as well as (3) the joint model trained to optimise both.

The models are based on DistilBERT \cite{sanh2019distilbert}, which is a reduced version of BERT \cite{devlin2019bert} performing on par with it as reported by the authors. For each of the models described below, we take the version of DistilBERT that is pre-trained with a language-modelling objective and further fine-tune its embeddings for the specific task at hand. 

\subsection{Generating Explanations}\label{sec:explanationGen}

\begin{figure*}[t]
\centering
\includegraphics[width=\linewidth]{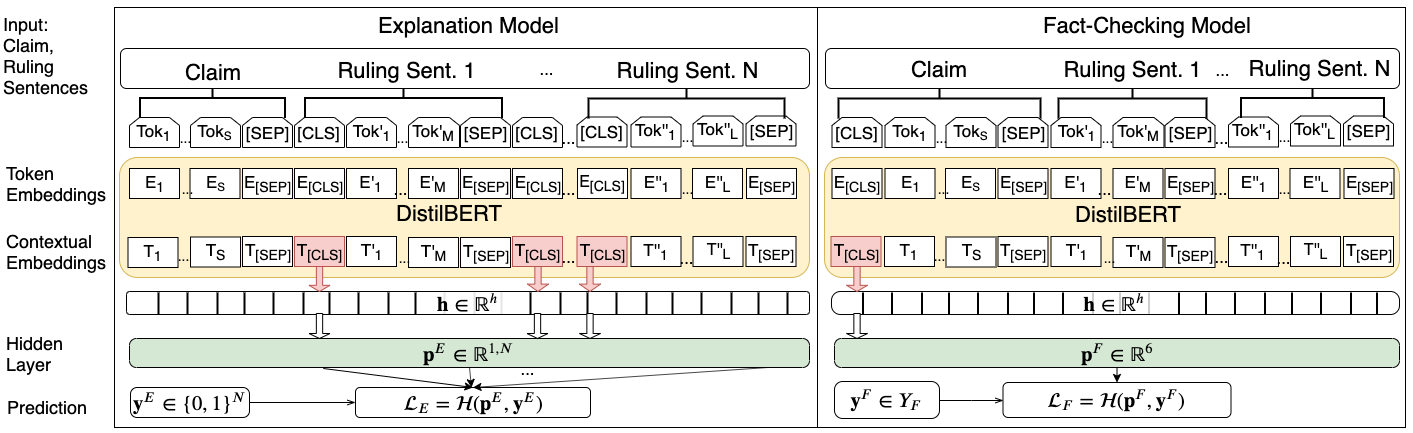}
\caption{Architecture of the \textit{Explanation} (left) and \textit{Fact-Checking} (right) models that optimise separate objectives.} 
\label{figure:separateModels}
\end{figure*}

Our explanation model, shown in Figure~\ref{figure:separateModels} (left) is inspired by the recent success of utilising the transformer model architecture for extractive summarisation \cite{liu-lapata-2019-text}. It learns to maximize the similarity of the extracted explanation with the human justification.  

We start by greedily selecting the top $k$ sentences from each claim's ruling comments that achieve the highest ROUGE-2 F1 score when compared to the gold justification. We choose $k = 4$, as that is the average number of sentences in veracity justifications. The selected sentences, referred to as oracles, serve as positive gold labels - $\mathbf{y}^E \in \{0,1\}^N $, where $N$ is the total number of sentences present in the ruling comments. Appendix~\ref{appendix:a} provides an overview of the coverage that the extracted oracles achieve compared to the gold justification. Appendix~\ref{appendix:o} further presents examples of the selected oracles, compared to the gold justification. 

At training time, we learn a function $f(X) = \mathbf{p}^E$, $\mathbf{p}^E \in \mathbb{R}^{1, N}$ that, based on the input $X$, the text of the claim and the ruling comments, predicts which sentence should be selected - \{0,1\}, to constitute the explanation. At inference time, we select the top $n = 4$ sentences with the highest confidence scores.

Our extraction model, represented by function $f(X)$, takes the contextual representations produced by the last layer of DistilBERT and feeds them into a feed-forward task-specific layer - $\mathbf{h} \in \mathbb{R}^{h}$. It is followed by the prediction layer $\mathbf{p}^{E} \in \mathbb{R}^{1,N}$ with sigmoid activation. The prediction is used  to optimise the cross-entropy loss function $\mathcal{L}_{E}=\mathcal{H}(\mathbf{p}^{E}, \mathbf{y}^{E})$.

\subsection{Veracity Prediction}\label{sec:veracityPred}
For the veracity prediction model, shown in Figure~\ref{figure:separateModels} (right), we learn a function $g(X) = \mathbf{p}^F$ that, based on the input X, predicts the veracity of the claim $\mathbf{y}^{F} \in Y_{F}$, $Y_F =$ \textit{\{true, false, half-true, barely-true, mostly-true, pants-on-fire\}}. 

The function $g(X)$ takes the contextual token representations from the last layer of DistilBERT and feeds them to a task-specific feed-forward layer $\mathbf{h} \in \mathbb{R}^{h}$. It is followed by the prediction layer with a softmax activation $\mathbf{p}^{F} \in \mathbb{R}^{6}$. We use the prediction to optimise a cross-entropy loss function $\mathcal{L}_{F}= \mathcal{H}(\mathbf{p}^{F}, \mathbf{y}^{F})$.

\subsection{Joint Training}\label{sec:jointTraining}
\begin{figure}[t]
\centering
\includegraphics[width=180pt]{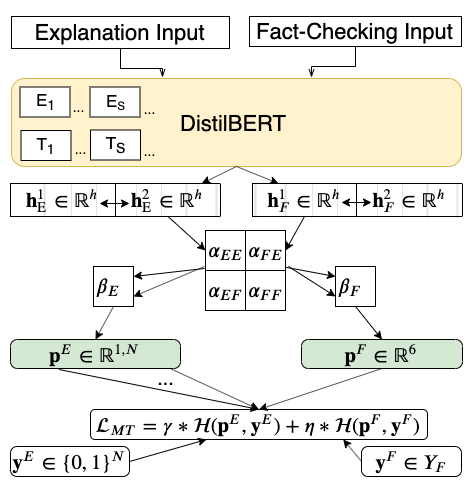}
\caption{Architecture of the \textit{Joint} model learning Explanation (E) and Fact-Checking (F) at the same time.}
\label{figure:jointmodel} 
\end{figure}

Finally, we learn a function $h(X) = (\mathbf{p}^E, \mathbf{p}^F)$ that, given the input X - the text of the claim and the ruling comments, predicts both the veracity explanation $\mathbf{p}^E$ and the veracity label $\mathbf{p}^F$ of a claim. The model is shown Figure~\ref{figure:jointmodel}. The function $h(X)$ takes the contextual embeddings $\mathbf{c}^E$ and $\mathbf{c}^F$ produced by the last layer of DistilBERT and feeds them into a cross-stitch layer \cite{misra2016cross,ruder122019latent}, which consists of two layers with two shared subspaces each - $\mathbf{h}_{E}^1$ and $\mathbf{h}_{E}^2$ for the explanation task and $\mathbf{h}_F^1$ and $\mathbf{h}_F^2$ for the veracity prediction task. In each of the two layers, there is one subspace for task-specific representations and one that learns cross-task representations. The subspaces and layers interact trough $\alpha$ values, creating the linear combinations $\widetilde{h}^i_E$ and $\widetilde{h}^j_F$, where i,j$\in \{1,2\}$:
%
\begin{equation}
\centering
\begin{bmatrix}
\widetilde{h}^i_E\\ 
\widetilde{h}^j_F
\end{bmatrix}
=
\begin{bmatrix}
\alpha_{EE} & \alpha_{EF}\\ 
\alpha_{FE} & \alpha_{FF}
\end{bmatrix}
\begin{bmatrix}
{h^i_E}^T & {h^j_F}^T\\ 
\end{bmatrix}  
\end{equation}

We further combine the resulting two subspaces for each task - $\widetilde{h}^i_E$ and $\widetilde{h}^j_F$ with parameters $\beta$ to produce one representation per task:
\begin{equation}
\centering
\widetilde{h}^T_P
=
\begin{bmatrix}
\beta_P^1\\ 
\beta_P^2
\end{bmatrix}^T
\begin{bmatrix}
\widetilde{h}^1_P & \widetilde{h}^2_P\\ 
\end{bmatrix}^T
\end{equation}
where P $\in \{E, F\}$ is the corresponding task.

Finally, we use the produced representation to predict $\mathbf{p}^{E}$ and $\mathbf{p}^{F}$, with feed-forward layers followed by sigmoid and softmax activations accordingly. We use the prediction to optimise the joint loss function $\mathcal{L}_{MT}= \gamma*\mathcal{H}(\mathbf{p}^{E}, \mathbf{y}^{E}) + \eta * \mathcal{H}(\mathbf{p}^{F}, \mathbf{y}^{F})$, where $\gamma$ and $\eta$ are used for weighted combination of the individual loss functions.

\section{Automatic Evaluation}\label{sec:automaticEval}
We first conduct an automatic evaluation of both the veracity prediction and veracity explanation models. 

\subsection{Experiments}\label{subsec:automaticExperiments}
In Table~\ref{tab:results:explanation}, we compare the performance of the two proposed models for generating extractive explanations. \textit{Explain-MT} is trained jointly with a veracity prediction model, and \textit{Explain-Extractive} is trained separately. We include the \textit{Lead-4} system \cite{Nallapati:2017:SRN:3298483.3298681} as a baseline, which selects as a summary the first four sentences from the ruling comments. The \textit{Oracle} system presents the best greedy approximation of the justification with sentences extracted from the ruling comments. It indicates the upper bound that could be achieved by extracting sentences from the ruling comments as an explanation. The performance of the models is measured using ROUGE-1, ROUGE-2, and ROUGE-L F1 scores.

In Table~\ref{tab:results:fact-checking}, we again compare two models - one trained jointly - \textit{MT-Veracity@Rul}, with the explanation generation task and one trained separately - \textit{Veracity@Rul}. As a baseline, we report the work of \citet{wang2017liar}, who train a model based on the metadata available about the claim. It is the best known model that uses only the information available from the LIAR dataset and not the gold justification, which we aim at generating. 

We also provide two upper bounds serving as an indication of the approximate best performance that can be achieved given the gold justification. The first is the reported system performance from \citet{alhindi-etal-2018-evidence}, and the second - \textit{Veracity@Just}, is our veracity prediction model but trained on gold justifications. The \citet{alhindi-etal-2018-evidence} system is trained using a BiLSTM, while we train the \textit{Veracity@Just} model using the same model architecture as for predicting the veracity from the ruling comments with \textit{Veracity@Rul}. 

Lastly, \textit{Veracity@RulOracles} is the veracity model trained on the gold oracle sentences from the ruling comments. It provides a rough estimate of how much of the important information from the ruling comments is preserved in the oracles. The models are evaluated with a macro F1 score.

\subsection{Experimental Setup}\label{sec:experiments}

Our models employ the base, uncased version of the pre-trained DistilBERT model. The models are fed with text depending on the task set-up - claim and ruling sentences for the explanation and joint models; claim and ruling sentences, claim and oracle sentences or claim and justification for the fact-checking model. We insert a `[CLS]' token before the start of each ruling sentence (explanation model), before the claim (fact-checking model), or at the combination of both for the joint model. The text sequence is passed through a number of Transformer layers from DistilBERT. We use the `[CLS]' embeddings from the final contextual layer of DistilBERT and feed that in task-specific feed-forward layers $\mathbf{h} \in \mathbb{R}^{h}$, where h is 100 for the explanation task, 150 for the veracity prediction one and 100 for each of the joint cross-stitch subspaces. Following are the task-specific prediction layers ${p}^E$. 

The size of $h$ is picked with grid-search over \{50, 100, 150, 200, 300\}. We also experimented with replacing the feed-forward task-specific layers with an RNN or Transformer layer or including an activation function, which did not improve task performance.

The models are trained for up to 3 epochs, and, following \citet{liu-lapata-2019-text}, we evaluate the performance of the fine-tuned model on the validation set at every 50 steps, after the first epoch. We then select the model with the best ROUGE-2 F1 score on the validation set, thus, performing a potential early stopping. The learning rate used is 3e-5, which is chosen with a grid search over \{3e-5, 4e-5, 5e-5\}. We perform 175 warm-up steps (5\% of the total number of steps), after also experimenting with 0, 100, and 1000 warm-up steps. Optimisation is performed with AdamW \cite{loshchilov2017fixing}, and the learning rate is scheduled with a warm-up linear schedule \cite{goyal2017accurate}. The batch size during training and evaluation is 8.

The maximum input words to DistilBERT are 512, while the average length of the ruling comments is 904 words. To prevent the loss of any sentences from the ruling comments, we apply a sliding window over the input of the text and then merge the contextual representations of the separate sliding windows, mean averaging the representations in the overlap of the windows. The size of the sliding window is 300, with a stride of 60 tokens, which is the number of overlapping tokens between two successive windows. The maximum length of the encoded sequence is 1200. We find that these hyper-parameters have the best performance after experimenting with different values in a grid search.

We also include a dropout layer (with 0.1 rate for the separate and 0.15 for the joint model) after the contextual embedding provided by the transformer models and after the first linear layer as well.

The models optimise cross-entropy loss, and the joint model optimises a weighted combination of both losses. Weights are selected with a grid search - 0.9 for the task of explanation generation and 0.1 for veracity prediction. The best performance is reached with weights that bring the losses of the individual models to roughly the same scale.

\subsection{Results and Discussion}\label{subsec:automaticResults}

\begin{table}
\fontsize{10}{10}\selectfont
\centering
\begin{tabular}{lll}
\toprule
\textbf{Model} & \textbf{Val} & \textbf{Test}  \\ 
\midrule
\citet{wang2017liar}, all metadata & 0.247 & 0.274 \\
\midrule
Veracity@RulOracles & 0.308 & 0.300 \\ 
Veracity@Rul & 0.313 & 0.313 \\ 
MT-Veracity@Rul & \textbf{0.321} & \textbf{0.323}  \\ 
\midrule
\citet{alhindi-etal-2018-evidence}@Just & 0.37 & 0.37 \\ 
Veracity@Just & \textbf{0.443}& \textbf{0.443} \\ 
\bottomrule
\end{tabular}
\caption{Results (Macro F1 scores) of the veracity prediction task on all of the six classes. The models are trained using the text from the ruling oracles (@RulOracles), ruling comment (@Rul), or the gold justification (@Just).}
\label{tab:results:fact-checking}
\end{table}

\begin{table*}[h!]
\fontsize{10}{10}\selectfont
\centering
\begin{tabular}{l|ccc|ccc}
\toprule
\multirow{2}{*}{\textbf{Model}} & \multicolumn{3}{c|}{\textbf{Validation}}&  \multicolumn{3}{c}{\textbf{Test}} \\ 
& \textbf{ROUGE-1} & \textbf{ROUGE-2} & \textbf{ROUGE-L} & \textbf{ROUGE-1} & \textbf{ROUGE-2} & \textbf{ROUGE-L} \\ \midrule
Lead-4 & 27.92 & 6.94 & 24.26 & 28.11 & 6.96 & 24.38 \\
Oracle & 43.27 & 22.01 & 38.89 & 43.57 & 22.23 & 39.26 \\
\midrule
Explain-Extractive & \textbf{35.64} & \textbf{13.50} & \textbf{31.44} & \textbf{35.70} & \textbf{13.51} & \textbf{31.58} \\
Explain-MT & 35.18 & 12.94 & 30.95 & 35.13 & 12.90 & 30.93 \\
\bottomrule 		
\end{tabular}
\caption{Results of the veracity explanation generation task. The results are ROUGE-N F1 scores of the gene- \newline rated explanation w.r.t. the gold justification.} 
\label{tab:results:explanation}
\end{table*} 

For each claim, our proposed joint model (see \ref{sec:jointTraining}) provides both (i) a veracity explanation and (ii) a veracity prediction. We compare our model's performance with models that learn to optimise these objectives \emph{separately}, as no other joint models have been proposed.
Table~\ref{tab:results:fact-checking} shows the results of veracity prediction, measured in terms of macro F1. 

Judging from the performance of both \textit{Veracity@Rul} and \textit{MT-Veracity@Rul}, we can assume that the task is very challenging. Even given a gold explanation (\citet{alhindi-etal-2018-evidence} and \textit{Veracity@Just}), the macro F1 remains below 0.5. This can be due to the small size of the dataset and/or the difficulty of the task even for human annotators. We further investigate the difficulty of the task in a human evaluation, presented in Section~\ref{sec:manualEval}. 


Comparing \textit{Veracity@RulOracles} and \textit{Veracity@Rul}, the latter achieves a slightly higher macro F1 score, indicating that the extracted ruling oracles, while approximating the gold justification, omit information that is important for veracity prediction. Finally, when the fact checking system is learned jointly with the veracity explanation system - \textit{MT-Veracity@Rul}, it achieves the best macro F1 score of the three systems. The objective to extract explanations provides information about regions in the ruling comments that are close to the gold explanation, which helps the veracity prediction model to choose the correct piece of evidence.

In Table~\ref{tab:results:explanation}, we present an evaluation of the generated explanations, computing ROUGE F1 score w.r.t. gold justification. 
Our first model, the \textit{Explain-Extractive} system, optimises the single objective of selecting explanation sentences. It outperforms the baseline, indicating that generating veracity explanations is possible.

\textit{Explain-Extractive} also outperforms the \textit{Explain-MT} system. While we would expect that training jointly with a veracity prediction objective would improve the performance of the explanation model, as it does for the veracity prediction model, we observe the opposite. This indicates a potential mismatch between the ruling oracles and the salient regions for the fact checking model. We also find a potential indication of that in the observed performance decrease when the veracity model is trained solely on the ruling oracles compared to the one trained on all of the ruling comments. We hypothesise that, when trained jointly with the veracity extraction component, the explanation model starts to also take into account the actual knowledge needed to perform the fact check, which might not match the exact wording present in the oracles, thus decreasing the overall performance of the explanation system. We further investigate this in a manual evaluation of which of the systems - Explain-MT and Explain-Extractive, generates explanations with better qualities and with more information about the veracity label.

Finally, comparing the performance of the extractive models and the \textit{Oracle}, we can conclude that there is still room for improvement of explanation systems when only considering extractive summarisation.

\subsection{A Case Study}\label{section:case}
Table~\ref{tab:example} presents two example explanations generated by the extractive vs. the multi-task model.
In the first example, the multi-task explanation achieves higher ROUGE scores than the extractive one. The corresponding extractive summary contains information that is not important for the final veracity label, which also appears to affect the ROUGE scores of the explanation. On the other hand, the multi-task model, trained jointly with a veracity prediction component, selects sentences that are more important for the fact check, which in this case is also beneficial for the final ROUGE score of the explanation.

In the second example, the multi-task explanation has lower ROUGE scores than the extractive one. We observe that the gold justification contains some sentences that are not relevant to the fact check, and the extractive summary is fooled to select explanation sentences that are close to the gold summary. As a result, the explanation does not provide enough information about the chosen veracity label. The multi-task model, on the other hand, selects sentences that are also contributing to the prediction of the veracity labels. Thus, its explanation turns out to be more beneficial for the final fact check even though it has a lower ROUGE score compared to the gold justification.

\begin{table*}
\centering
\scriptsize
\begin{tabular}{p{\linewidth}}
\toprule

\textbf{Label: }Mostly-true \textbf{Claim:} Says opponent U.S. Rep. Gwen Moore is Wisconsin's most absent member of Congress, missing nearly 17 percent of the House votes in the second quarter of 2012. \\
\textbf{Just:} Sebring says Moore is the state's "most absent" member of Congress and that she missed 17 percent of the House votes in the second quarter of 2012. Indeed, Moore missed nearly 17 percent of the votes in the second quarter. That is Moore's worst quarter, but it also amounts to Sebring doing some cherry-picking of data. A longer period shows a lower percentage, but even that places Moore at the top of the state's eight-member delegation. \\
\textbf{Explain-Extr:} Then he turns to the ad's main point: "Gwen Moore is Wisconsin's most absent member of Congress, missing nearly 17 percent of the House votes in the second quarter of 2012." That, again, compares to the median of 2.5 percent, according to GovTrack. Overall, the state delegation missed an average of 2.53 percent of the votes this session. But what about the other part of Sebring's claim, that Moore is the "most absent" member of the state delegation? \\
\textbf{Explain-MT:} Then he turns to the ad's main point: "Gwen Moore is Wisconsin's most absent member of Congress, missing nearly 17 percent of the House votes in the second quarter of 2012." If you take out the first two quarters of 2012, Moore missed an average of 3.4 percent of the votes in the other five quarters of this session. In the second quarter, the one cited by Sebring, the missed votes easily put Moore at the top of the state's eight-member House delegation. Looking at the full session, Moore missed 112 votes, or 6.98 percent. \\
\midrule
\textbf{Label:} Half-true \textbf{Claim:} Of the more than 1.3 million temporary mortgage modifications, over half have defaulted. \\
\textbf{Just:} In the final full week of the U.S. Senate race, how did Rubio fare on his numbers about the mortgage modification program? Rubio said "over 1.3 million temporary work-outs, over half have now defaulted,'' referring to a temporary mortgage modification program. \\
\textbf{Explain-Extr:} Over 1.3 million temporary work-outs, over half have now defaulted,'' he said. There have been 500,000 permanent (modifications). Rubio also said that more than half of those 1.3 million had defaulted." Rubio: "The temporary modifications." \\
\textbf{Explain-MT:} Rubio also said that more than half of those 1.3 million had "defaulted." "Over 1.3 million temporary work-outs, over half have now defaulted,'' he said. Of those permanent modifications, the majority survived while almost 29,000 were cancelled. The chart states that 699,924 trials had been cancelled -- so note that is slightly more than half.\\
\bottomrule
\end{tabular}
\caption{Examples of the generated explanation of the extractive (Explain-Extr) and the multi-task model (Explain-MT) compared to the gold justification (Just).}
\label{tab:example}
\end{table*}

\section{Manual Evaluation}\label{sec:manualEval}
As the ROUGE score only accounts for word-level similarity between gold and predicted justifications, we also conduct a manual evaluation of the quality of the produced veracity explanations.

\subsection{Experiments}\label{subsec:manualExperiments}
\textbf{Explanation Quality}. We first provide a manual evaluation of the properties of three different types of explanations - gold justification, veracity explanation generated by the  \textit{Explain-MT}, and the ones generated by \textit{Explain-Extractive}. We ask three annotators to rank these explanations with the ranks 1, 2, 3, (first, second, and third place) according to four different criteria:

\begin{enumerate}[noitemsep]
    \item \textbf{Coverage.} The explanation contains important, salient information and does not miss any important points that contribute to the fact check.
    \item \textbf{Non-redundancy.} The summary does not contain any information that is redundant/repeated/not relevant to the claim and the fact check.
    \item \textbf{Non-contradiction.} The summary does not contain any pieces of information that are contradictory to the claim and the fact check. 
    \item \textbf{Overall.} Rank the explanations by their overall quality.
\end{enumerate}

We also allow ties, meaning that two veracity explanations can receive the same rank if they appear the same. 

For the annotation task set-up, we randomly select a small set of 40 instances from the test set and collect the three different veracity explanations for each of them. We did not provide the participants with information of the three different explanations and shuffled them randomly to prevent easily creating a position bias for the explanations. The annotators worked separately without discussing any details about the annotation task.

\textbf{Explanation Informativeness}. In the second manual evaluation task, we study how well the veracity explanations manage to address the information need of the user and if they sufficiently describe the veracity label. We, therefore, design the annotation task asking annotators to provide a veracity label for a claim based on a veracity explanation coming from the justification, the \textit{Explain-MT}, or the \textit{Explain-Extractive} system. The annotators have to provide a veracity label on two levels - binary classification - true or false, and six-class classification - true, false, half-true, barely-true, mostly-true, pants-on-fire. Each of them has to provide the label for 80 explanations, and there are two annotators per explanation. 

\subsection{Results and Discussion}\label{sec:manualResults}
\textbf{Explanation Quality}. Table~\ref{tab:results:man1} presents the results from the manual evaluation in the first set-up, described in Section~\ref{sec:manualEval}, where annotators ranked the explanations according to four different criteria. 

We compute Krippendorff's $\alpha$ inter-annotator agreement (IAA, \citet{hayes2007answering}) as it is suited for ordinal values. The corresponding alpha values are 0.26 for \textit{Coverage}, 0.18 for \textit{Non-redundancy}, -0.1 for \textit{Non-contradiction}, and 0.32 for \textit{Overall}, where $0.67<\alpha <0.8$ is regarded as significant, but vary a lot for different domains. 

We assume that the low IAA can be attributed to the fact that in ranking/comparison tasks for manual evaluation, the agreement between annotators might be affected by small differences in one rank position in one of the annotators as well as by the annotator bias towards ranking explanations as ties. Taking this into account, we choose to present the mean average recall for each of the annotators instead. Still, we find that their preferences are not in a perfect agreement and report only what the majority agrees upon. We also consider that the low IAA reveals that the task might be ``already too difficult for humans''. This insight proves to be important on its own as existing machine summarisation/question answering studies involving human evaluation do not report IAA scores \cite{liu-lapata-2019-text}, thus, leaving essential details about the nature of the evaluation tasks ambiguous. 


We find that the gold explanation is ranked the best for all criteria except for \textit{Non-contradiction}, where one of the annotators found that it contained more contradictory information than the automatically generated explanations, but Krippendorff's $\alpha$ indicates that there is no agreement between the annotations for this criterion. 

Out of the two extractive explanation systems, \textit{Explain-MT} ranks best in Coverage and Overall criteria, with 0.21 and 0.13 corresponding improvements in the ranking position. These results contradict the automatic evaluation in Section~\ref{subsec:automaticResults}, where the explanation of \textit{Explain-MT} had lower ROUGE F1 scores. This indicates that an automatic evaluation might be insufficient in estimating the information conveyed by the particular explanation.

On the other hand, \textit{Explain-Extr} is ranked higher than \textit{Explain-MT} in terms of Non-redundancy and Non-contradiction, where the last criterion was disagreed upon, and the rank improvement for the first one is only marginal at 0.04. 

This implies that a veracity prediction objective is not necessary to produce natural-sounding explanations (\textit{Explain-Extr}), but that the latter is useful for generating better explanations overall and with higher coverage \textit{Explain-MT}. 

\begin{table}[h!]
\fontsize{10}{10}\selectfont
\centering
\begin{tabular}{lccc}
\toprule
\textbf{Annotators} & \textbf{Just} & \textbf{Explain-Extr} & \textbf{Explain-MT}\\ \midrule
\multicolumn{4}{c}{Coverage} \\ \midrule
All & \textbf{1.48} & 1.89 & \cellcolor{myblue}1.68 \\
1st & \textbf{1.50} & 2.08 & \cellcolor{myblue}1.87 \\
2nd & \textbf{1.74} & 2.16 & \cellcolor{myblue}1.84 \\
3rd & \textbf{1.21} & 1.42 & \cellcolor{myblue}1.34 \\ \midrule
\multicolumn{4}{c}{Non-redundancy} \\ \midrule
All & \textbf{1.48} & \cellcolor{myblue}1.75 & 1.79 \\
1st & \textbf{1.34} & 1.84 & \cellcolor{myblue}1.76 \\
2nd & \textbf{1.71} & \cellcolor{myblue}1.97 & 2.08 \\
3rd & \textbf{1.40} & \cellcolor{myblue}1.42 & 1.53 \\ \midrule
\multicolumn{4}{c}{Non-contradiction} \\ \midrule
All & 1.45 & \cellcolor{myblue}\textbf{1.40} & 1.48 \\
1st & \textbf{1.13} & 1.45 & \cellcolor{myblue}1.34 \\
2nd & 2.18  & \cellcolor{myblue}\textbf{1.63} & 1.92 \\
3rd & \textbf{1.03} & \cellcolor{myblue}1.13  & 1.18 \\ \midrule
\multicolumn{4}{c}{Overall} \\ \midrule
All & \textbf{1.58} & 2.03 & \cellcolor{myblue}1.90 \\
1st & \textbf{1.58} & 2.18 & \cellcolor{myblue}1.95 \\
2nd & \textbf{1.74} & 2.13 & \cellcolor{myblue}1.92 \\
3rd & \textbf{1.42} & \cellcolor{myblue}1.76  & 1.82 \\
\bottomrule 		
\end{tabular}

\caption{Mean Average Ranks (MAR) of the explanations for each of the four evaluation criteria. The explanations come from the gold justification (Just), the generated explanation (Explain-Extr), and the explanation learned jointly (Explain-MT) with the veracity prediction model. The lower MAR indicates a higher ranking, i.e., a better quality of an explanation. For each row, the best results are in bold, and the best results with automatically generated explanations are in blue.} 
\label{tab:results:man1}
\end{table} 

\textbf{Explanation Informativeness}. Table~\ref{tab:results:man2} presents the results from the second manual evaluation task, where annotators provided the veracity of a claim based on an explanation from one of the systems. We here show the results for binary labels, as annotators struggled to distinguish between 6 labels. 
The latter follows the same trends and are shown in Appendix~\ref{appendix:q}. 

The Fleiss' $\kappa$ IAA for binary prediction is: \textit{Just} -- 0.269,  \textit{Explain-MT} -- 0.345, \textit{Explain-Extr} -- 0.399. The highest agreement is achieved for \textit{Explain-Extr}, which is supported by the highest proportion of agreeing annotations from Table~\ref{tab:results:man2}. Surprisingly, the gold explanations from \textit{Just} were most disagreed upon. 
Apart from that, looking at the agreeing annotations, gold explanations were found most sufficient in providing information about the veracity label and also were found to explain the correct label most of the time. They are followed by the explanations produced by \textit{Explain-MT}. This supports the findings of the first manual evaluation, where the \textit{Explain-MT} ranked better in coverage and overall quality than \textit{Explain-Extr}.

\begin{table}[t]
\fontsize{10}{10}\selectfont
\centering
\begin{tabular}{clrrr}
\toprule
 & & \textbf{Just} & \textbf{Explain-Extr} & \textbf{Explain-MT} \\ \midrule
$\nwarrow$ & Agree-C & \textbf{0.403} & 0.237 & \cellcolor{myblue}0.300 \\
$\searrow$ & Agree-NS & \textbf{0.065} &  0.250 & \cellcolor{myblue}0.188 \\
$\searrow$ & Agree-NC & \textbf{0.064} & 0.113 & \cellcolor{myblue}0.088 \\
$\searrow$ & Disagree & 0.468 & \cellcolor{myblue}\textbf{0.400} & 0.425\\
\bottomrule 		
\end{tabular}
\caption{Manual veracity labelling, given a particular explanation from the gold justification (Just), the generated explanation (Explain-Extr), and the explanation learned jointly (Explain-MT) with the veracity prediction model. Percentages of the dis/agreeing annotator predictions are shown, with agreement percentages split into: \emph{correct} according to the gold label (Agree-C), \emph{incorrect} (Agree-NC) or \emph{insufficient information} (Agree-NS). The first column indicates whether higher ($\nwarrow$) or lower ($\searrow$) values are better. For each row, the best results are in bold, and the best results with automatically generated explanations are in blue.}
\label{tab:results:man2}
\end{table} 

\section{Related Work}
 
\textbf{Generating Explanations.}
Generating textual explanations for model predictions is an understudied problem. The first study was \citet{NIPS2018_8163}, who generate explanations for the task of natural language inference. The authors explore three different set-ups: prediction pipelines with explanation followed by prediction, and prediction followed by explanation, and a joint multi-task learning setting. They find that first generating the explanation produces better results for the explanation task, but harms classification accuracy. 

We are the first to provide a study on generating veracity explanations. We show that the generated explanations improve veracity prediction performance, and find that jointly optimising the veracity explanation and veracity prediction objectives improves the coverage and the overall quality of the explanations.

\textbf{Fact Checking Interpretability.} Interpreting fact checking systems has been explored in a few studies. \citet{shu2019defend} study the interpretability of a system that fact checks full-length news pages by leveraging user comments from social platforms. They propose a co-attention framework, which selects both salient user comments and salient sentences from news articles. \citet{yang2019xfake} build an interpretable fact-checking system XFake, where shallow student and self-attention, among others, are used to highlight parts of the input. This is done solely based on the statement without considering any supporting facts. 
In our work, we research models that generate human-readable explanations, and directly optimise the quality of the produced explanations instead of using attention weights as a proxy. We use the LIAR dataset to train such models, which contains fact checked single-sentence claims that already contain professional justifications.
As a result, we make an initial step towards automating the generation of professional fact checking justifications.
 
\textbf{Veracity Prediction.}
Several studies have built fact checking systems for the LIAR dataset \cite{wang2017liar}. The model proposed by \citet{karimi-etal-2018-multi} reaches 0.39 accuracy by using metadata, ruling comments, and justifications. \citet{alhindi-etal-2018-evidence} also trains a classifier, that, based on the statement and the justification, achieves 0.37 accuracy. To the best of our knowledge, \citet{long2017fake} is the only system that, without using justifications, achieves a performance above the baseline of \citet{wang2017liar}, an accuracy of 0.415---the current state-of-the-art performance on the LIAR dataset. Their model learns a veracity classifier with speaker profiles. 
While using metadata and external speaker profiles might provide substantial information for fact checking, they also have the potential to introduce biases towards a certain party or a speaker. 

In this study, we propose a method to generate veracity explanations that would explain the reasons behind a certain veracity label independently of the speaker profile. 
Once trained, such methods could then be applied to other fact checking instances without human-provided explanations or even to perform end-to-end veracity prediction and veracity explanation generation given a claim.

Substantial research on fact checking methods exists for the FEVER dataset~\cite{Thorne18Fever}, which comprises rewritten claims from Wikipedia.
Systems typically perform document retrieval, evidence selection, and veracity prediction. Evidence selection is performed using keyword matching \cite{malon-2018-team,yoneda-etal-2018-ucl}, supervised learning \cite{hanselowski-etal-2018-ukp, chakrabarty-etal-2018-robust} or sentence similarity scoring \cite{Ma:2018:DRS:3184558.3188729, mohtarami-etal-2018-automatic, Xu2019AdversarialDA}. More recently, the multi-domain dataset MultiFC \cite{augenstein-etal-2019-multifc} has been proposed, which is also distributed with evidence pages. Unlike FEVER, it contains real-world claims, crawled from different fact checking portals.

While FEVER and MultiFC are larger datasets for fact checking than LIAR-PLUS, they do not contain veracity explanations and can thus not easily be used to train joint veracity prediction and explanation generation models, hence we did not use them in this study.

\section{Conclusions}

We presented the first study on generating veracity explanations, and we showed that veracity prediction can be combined with veracity explanation generation and that the multi-task set-up improves the performance of the veracity system. A manual evaluation shows that the coverage and the overall quality of the explanation system is also improved in the multi-task set-up.

For future work, an obvious next step is to investigate the possibility of generating veracity explanations from evidence pages crawled from the Web. Furthermore, other approaches of generating veracity explanations should be investigated, especially as they could improve fluency or decrease the redundancy of the generated text.

\section{Acknowledgments}
\begin{wrapfigure}{L}{0.10\columnwidth}
\vspace{-13pt}
\includegraphics[width=0.17\columnwidth]{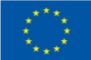}
\vspace{-25pt}
\end{wrapfigure}
This project has received funding from the European Union’s Horizon 2020 research and innovation programme under the Marie Skłodowska-Curie grant agreement No 801199.
\bibliography{anthology,acl2020}
\bibliographystyle{acl_natbib}

\clearpage
\appendix
\section{Appendices}\label{appendices}

\subsection{Comparison of different sources of evidence}\label{appendix:a}
Table \ref{tab:evidence} provides an overview of the ruling comments and the ruling oracles compared to the justification. The high recall in both ROUGE-1 and ROUGE-F achieved by the ruling comments indicates that there is a substantial coverage, i.e. over 70\% of the words and long sequences in the justification can be found in the ruling comments. On the other hand, there is a small coverage for the bi-grams. Selecting the oracles from all of the ruling sentences increases ROUGE-F1 scores mainly by improving the precision. 

\begin{table*}[h!] 
\centering
\begin{tabular}{l|rrr|rrr|rrr}
\toprule
\multirow{2}{*}{\textbf{Evidence Source}}& \multicolumn{3}{c|}{\textbf{ROUGE-1}}& \multicolumn{3}{c|}{\textbf{ROUGE-2}} & \multicolumn{3}{c}{\textbf{ROUGE-L}} \\ 
& P & R & F1 & P & R & F1 & P & R & F1 \\ \midrule
Ruling & 8.65 & 78.65 & 14.84 & 3.53 & 33.76 & 6.16 & 8.10 & 74.14 & 13.92 \\
Ruling Oracle & 43.97 & 49.24 & 43.79 & 22.45 & 24.50 & 22.03 & 39.70 & 44.10 & 39.37 \\ \bottomrule
\end{tabular}
\caption{Comparison of sources of evidence - Ruling Comments and Ruling Oracles comapred to the target \newline justification summary.}
\label{tab:evidence}
\end{table*}

\subsection{Extractive Gold Oracle Examples}
\label{appendix:o}
Table~\ref{tab:oracle-example} presents examples of selected oracles that serve as gold labels during training the extractive summarization model. The three examples represent oracles with different degrees of matching the gold summary. The first row presents an oracle that matches the gold summary with a ROUGE-L F1 score of 60.40 compared to the gold summary. It contains all of the important information from the gold summary and even points precise, not rounded, numbers. The next example has a ROUGE-L F1 score of 43.33, which is close to the average ROUGE-L F1 score for the oracles. The oracle again conveys the main points from the gold justification, thus, being sufficient for the claim's explanation. Finally, the third example is of an oracle with a ROUGE-L F1 score of 25.59. The selected oracle sentences still succeed in presenting the main points from the gold justification, which is at a more detailed level presenting specific findings. The latter might be found as a positive consequence as it presents the particular findings of the journalist that led to selecting the veracity label.

\begin{table*}
\centering
\scriptsize
\begin{tabular}{p{\linewidth}}
\toprule
\textbf{Claim: }``The president promised that if he spent money on a stimulus program that unemployment would go to 5.7 percent or 6 percent. Those were his words.'' \\
\textbf{Label: }Mostly-False  \\ 
\textbf{Just:} Bramnick said ``the president promised that if he spent money on a stimulus program that unemployment would go to 5.7 percent or 6 percent.
Those were his words.''
Two economic advisers estimated in a 2009 report that with the stimulus plan, the unemployment rate would peak near 8 percent before dropping to less than 6 percent by now.
Those are critical details Bramnick’s statement ignores.
To comment on this ruling, go to NJ.com. \\
\textbf{Oracle:}  ``The president promised that if he spent money on a stimulus program that unemployment would go to 5.7 percent or 6 percent.
Those were his words,'' Bramnick said in a Sept. 7 interview on NJToday.
But with the stimulus plan, the report projected the nation’s jobless rate would peak near 8 percent in 2009 before falling to about 5.5 percent by now.
So the estimates in the report were wrong.\\

\midrule
\textbf{Claim: }The Milwaukee County bus system has ``among the highest fares in the nation.''  \\
\textbf{Label:} False \\
\textbf{Just:} Larson said the Milwaukee County bus system has ``among the highest fares in the nation.''
But the system’s’ \$2.25 cash fare wasn’t at the top of a national comparison, with fares reaching as high as \$4 per trip.
And regular patrons who use a Smart Card are charged just \$1.75 a ride, making the Milwaukee County bus system about on par with average costs. \\
\textbf{Oracle:} Larson said the Milwaukee County bus system has ``among the highest fares in the nation.''
Patrons who get a Smart Card pay \$1.75 per ride.
At the time, nine cities on that list charged more than Milwaukee’s \$2.25 cash fare.
The highest fare -- in Nashville -- was \$4 per ride.\\
\midrule
 \textbf{Claim: }``The Republican who was just elected governor of the great state of Florida paid his campaign staffers, not with money, but with American Express gift cards.''  \\
 \textbf{Label: }Half-True \\
\textbf{Just:} First, we think many people might think Maddow was referring to all campaign workers, but traditional campaign staffers -- the people working day in and day out on the campaign -- were paid by check, like any normal job.
A Republican Party official said it was simply an easier, more efficient and quicker way to pay people.
And second, it's not that unusual.
In 2008, Obama did the same thing. \\
\textbf{Oracle:}
``It's a simpler and quicker way of compensating short-term help.''
Neither Conston nor Burgess said how many temporary campaign workers were paid in gift cards.
When asked how he was paid, Palecheck said: ``Paid by check, like any normal employee there.''
In fact, President Barack Obama's campaign did the same thing in 2008. \\
\bottomrule
\end{tabular}
\caption{Examples of the extracted oracle summaries (Oracle) compared to the gold justification (Just).}
\label{tab:oracle-example}
\end{table*}

\subsection{Manual 6-way Veracity Prediction from explanations}\label{appendix:q}
The Fleiss' $\kappa$ agreement for the 6-label manual annotations is: 0.20 on the \textit{Just} explanations, 0.230 on the \textit{Explain-MT} explanations, and 0.333 on the \textit{Explain-Extr} system. Table~\ref{tab:results:man2:6-way} represent the results of the manual veracity prediction with six classes.

\begin{table}[h!]
\centering
\begin{tabular}{clrrr}
\toprule
 & & Just & Explain-Extr & Explain-MT \\ \midrule
$\nwarrow$ & Agree-C & \textbf{0.208} & 0.138 & \cellcolor{myblue}0.163 \\
$\searrow$ & Agree-NS & \textbf{0.065} &  0.250 & \cellcolor{myblue}0.188 \\
$\searrow$ & Agree-NC & \textbf{0.052} & 0.100 & \cellcolor{myblue}0.075 \\
$\searrow$ & Disagree & 0.675 & \cellcolor{myblue}\textbf{0.513} & 0.575\\
\bottomrule 		
\end{tabular}
\caption{Manual classification of veracity label - true, false, half-true, barely-true, mostly-true, pants-on-fire, given a particular explanations from the gold justification (Just), the generated explanation (Explain-Extr) and the explanation learned jointly with the veracity prediction model (Explain-MT). Presented are percentages of the dis/agreeing annotator predictions, where the agreement percentages are split to: correct according to the gold label (Agree-C) , incorrect (Agree-NC) or with not sufficient information (Agree-NS). The first column indicates whether higher ($\nwarrow$) or lower ($\searrow$) values are better. At each row, the best set of explanations is in bold and the best automatic explanations are in blue.}
\label{tab:results:man2:6-way}
\end{table}


\end{document}